\documentclass[10pt,twocolumn,letterpaper]{article}
\usepackage{wacv}
\usepackage{booktabs}
\usepackage{amsmath}
\usepackage{amssymb}
\usepackage{graphicx}
\usepackage{multirow}
\usepackage{array}
\usepackage{float}  
\usepackage{placeins}  
\usepackage[dvipsnames]{xcolor}
\usepackage{balance}  
\usepackage{tikz}
\usetikzlibrary{arrows.meta,positioning,calc,fit,backgrounds}
\definecolor{cO}{HTML}{CBD5E1}
\definecolor{cB}{HTML}{4C9F70}
\definecolor{cI}{HTML}{5B8FB9}
\definecolor{cE}{HTML}{E08A3C}
\definecolor{cGrey}{HTML}{9AA3AC}

\usepackage{url}
\definecolor{wacvblue}{rgb}{0.21,0.49,0.74}
\usepackage[breaklinks,colorlinks,allcolors=wacvblue]{hyperref}

\usepackage{xspace}
\newcommand{\bioe}{BIOE\xspace}
\newcommand{\bioecv}{\textbf{BEACON}\xspace}
\newcommand{\bioecvr}{\textbf{BEACON}\xspace}
\newcommand{\mFoneB}{\ensuremath{\mathrm{mF1B}}\xspace}
\newcommand{\mFoneS}{\ensuremath{\mathrm{mF1S}}\xspace}
\newcommand{\wer}{\ensuremath{\mathrm{WER}}\xspace}

\def\confName{WACV}
\def\confYear{2027}

\begin{document}
\title{BEACON: Distilled Begin/End-Aware Structured Decoding for Continuous Sign Language Segmentation}

\author{Guohou Zhang$^{1}$ \qquad Kalin Stefanov$^{2}$\\
{\small $^{1}$Monash University, Australia \qquad
$^{2}$Japan Advanced Institute of Science and Technology, Japan}\\
{\tt\small gzha0067@student.monash.edu \qquad kalins@jaist.ac.jp}}

\maketitle

\begin{abstract}
In segmentation-based continuous sign language recognition, an untrimmed video is first partitioned into lexical units; segmentation is therefore a means to an end---downstream recognition.
However, segmenters are usually judged by local boundary detection rather than by the quality of the lexical segments and the recognition they enable.
We present \bioecvr, a begin/end-aware segmenter that combines four-class Outside/Begin/Inside/End (BIOE) labeling, constrained Viterbi decoding over eight legal transitions, and boundary knowledge distilled from a binary teacher into a shared student.
Under strict no-cut evaluation with validation-based model selection, \bioecvr improves segment-F1 by 14.58/14.59 points and boundary-F1 by 2.17/1.22 points, and reduces the downstream word error rate (WER) by 9.10/4.15 points relative to the state-of-the-art in the PHOENIX-14/BSL Corpus.
To examine which segmentation metric better reflects downstream recognition, we evaluated five distinct architecture families.
Segment-F1 aligns more consistently with lower WER than boundary-F1 in both corpora (Spearman $\rho=-1.0/-0.9$ versus $-0.9/-0.8$).
These results support evaluating sign language segmentation in terms of segment quality and downstream recognition utility rather than local boundary accuracy alone.
\end{abstract}

\section{Introduction}
\label{sec:introduction}

In segmentation-based recognition pipelines, untrimmed sign language videos are first partitioned into lexical units that can be passed to an isolated sign language recognizer.
Since signers produce fluid, co-articulated motions without explicit pauses, automatic segmentation remains an important bottleneck for pipelines that require localized lexical units; segmentation errors propagate into downstream recognition and contribute to a higher word error rate (\wer)~\cite{renz2021segmentation,moryossef2023seg}.
Segmentation is therefore a \emph{means to an end}, a perspective we make central both to how we \emph{evaluate} segmenters and to how we \emph{design} them.

\begin{figure}[t]
\centering
\includegraphics[width=\linewidth]{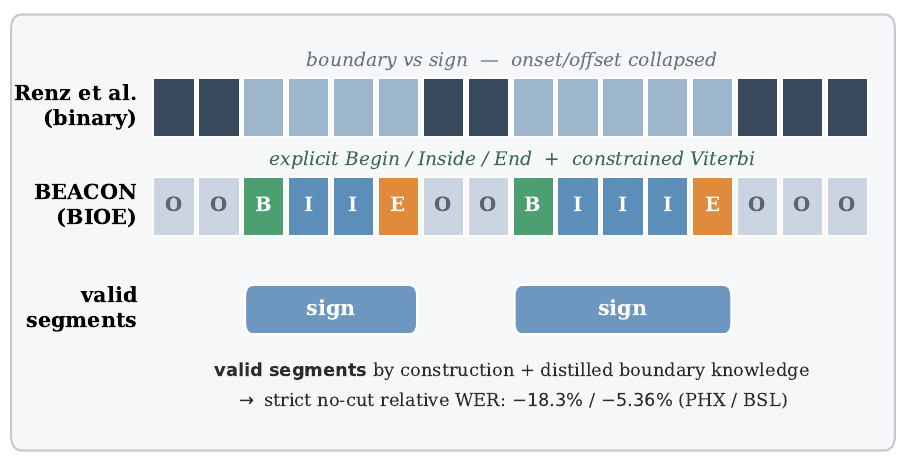}
\caption{\bioecv{} replaces Renz et al.~\cite{renz2021segmentation}'s binary boundary mask---which collapses sign onset/offset---with a 4-class \bioe{} labeling (explicit Outside/Begin/Inside/End states) and a constrained Viterbi decoder whose eight legal transitions make every output a \emph{well-formed segment sequence by construction}. This design improves segment quality while retaining boundary accuracy and lowers downstream recognition \wer.}
\label{fig:teaser}
\end{figure}

Renz et al.~\cite{renz2021segmentation} formulate segmentation as binary boundary detection: each frame is labeled either ``inside a sign'' or ``at a boundary''.
A multi-stage temporal convolutional network (MS-TCN)~\cite{farha2019mstcn} is trained with binary cross-entropy, followed by simple $\mathrm{argmax}$ decoding and an endpoint-cutting post-processing step.
The released evaluation procedure uses ground-truth evaluation endpoints for this operation; we revisit it in Sec.~\ref{sec:exp}.
This approach established a reproducible protocol on PHOENIX-14~\cite{koller2015phoenix14} and the BSL Corpus~\cite{schembri2013bsl} using boundary-F1 and segment-F1. Its binary representation, however, treats each frame only as either sign-internal or boundary, collapsing onset and offset into a single ``boundary'' class. It therefore does not explicitly represent how frames form a complete sign segment---where the segment begins, continues, and ends.

We make two departures.
First, we recast segmentation as \emph{structured sequence labeling}: every frame is tagged \texttt{O}/\texttt{B}/\texttt{I}/\texttt{E} (Outside/Begin/Inside/End), inspired by BIO chunk-tagging~\cite{ramshaw1995text}.
The scheme is binary-compatible---it maps exactly to Renz et al.~\cite{renz2021segmentation}'s labels, as verified on both corpora---while explicitly distinguishing sign onset and offset.
Fig.~\ref{fig:teaser} previews the central representational change from an undifferentiated boundary mask to a structurally valid Begin/Inside/End path.
These explicit states define a small finite-state grammar; constrained Viterbi decoding~\cite{viterbi1967error} over eight legal \bioe{} transitions restricts predictions to structurally valid paths and returns a well-formed segment sequence by construction.

Global path validity does not by itself provide sharp local endpoints.
A direct solution would be to pair the BIOE segmenter with a separate binary boundary model, but this would duplicate most of the temporal architecture.
\bioecvr{} instead combines structured segment modeling and boundary modeling within a single student through knowledge distillation~\cite{hinton2015distilling}.
A trained binary boundary model is frozen as a teacher, and its boundary information is distilled into a lightweight IO branch built on the shared student encoder.
The teacher is used only during training, leaving a single 5M-parameter student at inference for structured segment prediction and endpoint refinement.
The two components serve complementary purposes: the fixed grammar guarantees structurally valid output sequences, while boundary distillation provides local information for more precise endpoint refinement.
Matched ablations examine the contribution of each component.

Second, we ask a question the field has left implicit: \emph{Which metric should a segmenter optimize?}
Boundary-F1 (\mFoneB) is the de facto headline metric, but if segmentation serves downstream recognition, the more relevant target is the metric most closely aligned with \wer.
Under a matched frozen recognizer control, we compare five distinct architecture families on both PHOENIX-14~\cite{koller2015phoenix14} and the BSL Corpus~\cite{schembri2013bsl}.
Among the tested architecture families, segment-F1 (\mFoneS{}) orders downstream \wer more consistently than \mFoneB{} on both corpora.
We assess this claim through each metric's association with the independent downstream \wer{} measure, rather than assume that segment-level metrics are inherently preferable.
Accordingly, we evaluate \bioecvr{} not only by segmentation metrics, but also by whether its explicit segment structure and boundary-sensitive refinement translate into lower \wer{}.
Concretely, our contributions are threefold:

\begin{enumerate}
\item \textbf{Single structured segmenter with distilled boundary knowledge.}
\bioecvr{} combines four-class \bioe{} prediction, constrained decoding over eight legal transitions, and training-only boundary distillation, while requiring only a single 5M-parameter model at inference. Under strict no-cut evaluation, it improves \mFoneS{} by $14.58$/$14.59$ points and \mFoneB{} by $2.17$/$1.22$ points over Renz et al.~\cite{renz2021segmentation} on the PHOENIX-14/BSL Corpus.

\item \textbf{Recognition under strict no-cut protocol.}
With validation-based checkpoint and refinement-parameter selection, three seeds, and a frozen matched recognizer, \bioecvr{} without ground-truth endpoint cutting reduces \wer{} by $9.10$/$4.15$ points relative to Renz et al.~\cite{renz2021segmentation} on the PHOENIX-14/BSL Corpus.

\item \textbf{Architecture-level analysis of segmentation metrics and downstream recognition.}
Across five architecture families, \mFoneS{} orders downstream \wer{} more consistently than \mFoneB{} on both PHOENIX-14 and the BSL Corpus, providing an architecture-level test of which segmentation metric better reflects recognition utility.
\end{enumerate}


\section{Related Work}
\label{sec:related}

\subsection{Sign Language Segmentation}
\noindent \textbf{Directly comparable baselines.}
Renz et al.~\cite{renz2021segmentation} established the baseline protocol for this task using binary boundary masks, an MS-TCN~\cite{farha2019mstcn} decoded by $\mathrm{argmax}$, and an endpoint-cutting post-processing step that uses ground-truth endpoints.
Performance is evaluated with \mFoneB/\mFoneS{} on PHOENIX-14~\cite{koller2015phoenix14} and the BSL Corpus~\cite{schembri2013bsl}.
We use the same metrics, datasets, and I3D~\cite{carreira2017i3d} representations, while adopting validation-based model selection and strict no-cut endpoint protocol (see Sec.~\ref{sec:exp}).

Hands-On~\cite{low2025handson} is the closest recent baseline.
It also casts segmentation as sequence labeling, but uses three-class BIO with $\mathrm{argmax}$ decoding and stronger HaMeR~\cite{hamer} features on the BSL Corpus.
In contrast, \bioecvr{} uses an explicit \texttt{E} (End) class (four-class \bioe), replaces $\mathrm{argmax}$ with constrained Viterbi decoding~\cite{viterbi1967error} (valid segments by construction), and adds soft \texttt{B}/\texttt{E} auxiliary heads, a boundary-aware temporal smoothness loss, and boundary knowledge distillation into a shared student.
Since Hands-On uses HaMeR features, whereas our controlled setting uses the released I3D representations, we treat it as a feature-heterogeneous reference rather than a feature-matched comparison.

\noindent \textbf{Related but not directly comparable.}
Other segmentation work differs from our setting in dataset, supervision, or evaluation protocol.
CMPL~\cite{renz2021cmpl} uses changepoint-modulated pseudo-labeling with additional unlabeled target data in a source-free domain adaptation setting.
Since this differs from the fully supervised setting used by our method and primary baselines, we do not treat its reported results as a head-to-head comparison.

Belissen et al.~\cite{belissen2020dicta} released Dicta-Sign-LSF-v2 with a sliding-window-F1 baseline rather than the \mFoneB/\mFoneS{} protocol.
Moryossef et al.~\cite{moryossef2023seg} studied linguistically motivated segmentation of German Sign Language using per-document frame-F1/IoU.
Farfan et al.~\cite{farfan2025peru} applied MS-TCN and diffusion-based temporal segmentation to Peruvian Sign Language under a different dataset and evaluation setting.
Sim et al.~\cite{sim2024changepoint} performed unsupervised changepoint segmentation.
MHB~\cite{zhao2025mhb} incorporated handshape pre-training for boundary detection but evaluated on ASLLRP-S~\cite{asllrp} rather than the PHOENIX-14/BSL Corpus.

Since these studies differ in datasets, frame rates, annotations, supervision, or evaluation metrics, we do not treat them as directly comparable baselines.
Few prior methods are evaluated under the same \mFoneB/\mFoneS{} protocol.

The closest approaches are positioned in Table~\ref{tab:method-positioning} by learning setting, sequence modeling, and evaluation.
\begin{table}[H]
\centering
\caption{Positioning of the closest sign-segmentation approaches. DA and PL denote domain adaptation and pseudo-labeling. Matched \wer uses a fixed corpus-matched recognizer; Sec.~\ref{sec:results} reports our controlled re-evaluation.}
\label{tab:method-positioning}
\setlength{\tabcolsep}{1pt}
\renewcommand{\arraystretch}{1.15}
\footnotesize
\begin{tabular}{@{}l
>{\centering\arraybackslash}p{0.16\columnwidth}
>{\centering\arraybackslash}p{0.17\columnwidth}
>{\centering\arraybackslash}p{0.17\columnwidth}
>{\centering\arraybackslash}p{0.175\columnwidth}@{}}
\toprule
Property & Renz et al.~\cite{renz2021segmentation} & Hands-On~\cite{low2025handson} & CMPL~\cite{renz2021cmpl} & \bioecvr{} \\
\midrule
Setting & Supervised & Supervised & Source-free DA & Supervised \\
Input & I3D & HaMeR + 3D angles & I3D & I3D \\
Temporal model & MS-TCN & Transformer & MS-TCN & TCN + Transformer \\
Labels & Binary & BIO & Binary & BIOE + IO \\
Readout & $\mathrm{argmax}$ & $\mathrm{argmax}$ & $\mathrm{argmax}$ & Constrained Viterbi \\
Boundary cue & GT endpoint cut & --- & CP-modulated PL & Distilled IO \\
Matched \wer & No & No & No & Yes \\
\bottomrule
\end{tabular}
\end{table}

\subsection{Structured Prediction with Transition Graphs}

Structured sequence labeling commonly combines explicit label schemes such as BIO~\cite{ramshaw1995text} and BIOES~\cite{ratinov2009design,lample2016neural} with structured prediction models such as CRF~\cite{lafferty2001crf}.
Our \bioe{} scheme adapts BIOES by retaining explicit outside, begin, inside, and end states while omitting a separate singleton \texttt{S} state, yielding a compact four-state grammar with eight legal frame-to-frame transitions.

Because this label grammar is known a priori, we enforce these transitions directly with a non-learned constrained Viterbi decoder~\cite{viterbi1967error}, rather than learning transition scores from data.
This guarantees that the decoded \bioe{} path satisfies the declared segment grammar by construction.
We compare this fixed transition graph with a learned-transition CRF control in Sec.~\ref{sec:results-ablations}.

\subsection{Downstream Impact on Recognition}
Prior segmentation methods~\cite{renz2021segmentation,low2025handson,belissen2020dicta,zhao2025mhb} mostly stop at boundary- or segment-level scores; for example, MHB~\cite{zhao2025mhb} reports boundary detection on ASLLRP-S but does not report downstream recognition \wer under a matched recognizer.
To our knowledge, none of these methods reports downstream recognition \wer under matched recognizer control or examines which segmentation metric best aligns with it.
We close the loop on both corpora with a frozen per-corpus recognizer: a \emph{partial} continuous sign language recognition pipeline (segmenter followed by an isolated sign recognizer), not an end-to-end CTC~\cite{graves2006ctc} system.
Freezing the recognizer isolates the segmentation effect, and the five-architecture-family analysis tests whether \mFoneS{} orders \wer more consistently than \mFoneB{} (see Sec.~\ref{sec:results-families}).


\section{Method}
\label{sec:method}

\subsection{BIOE Labeling Scheme}
\begin{figure}[H]
\centering
\includegraphics[width=.98\linewidth,trim=0 10 0 2,clip]{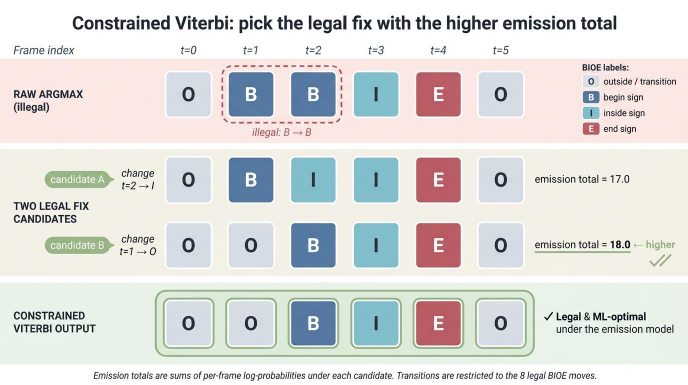}
\caption{Constrained Viterbi decoding~\cite{viterbi1967error} over the BIOE grammar. A framewise $\mathrm{argmax}$ path may contain illegal transitions; the decoder instead returns the highest emission path restricted to the eight legal BIOE moves, so the decoded sequence satisfies the declared grammar. Emission totals are illustrative.}
\label{fig:decoder}
\end{figure}

\begin{figure*}[t]
\centering
\includegraphics[width=.98\textwidth,trim=174 414 174 40,clip]{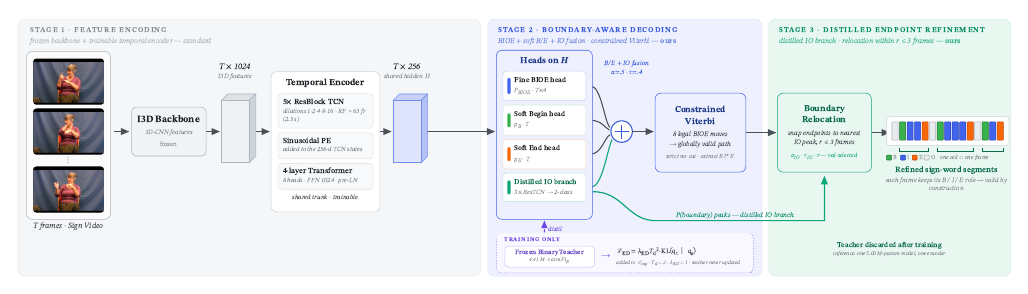}
\caption{The \bioecvr{} pipeline. A frozen I3D~\cite{carreira2017i3d} backbone supplies frame features to a shared temporal encoder (Stage 1, standard). BIOE and soft Begin/End predictions are fused and decoded over the eight legal BIOE transitions (inset), producing a preliminary segment path that satisfies the declared grammar (Stage 2, ours). A binary IO branch reads the same shared hidden states $H$ (no extra encoder) and refines endpoints within a validation-selected radius of at most $\pm3$ frames (Stage 3, ours). During training only, a separately trained, frozen binary teacher (shown dashed, with its own encoder, and discarded after training) distills boundary evidence into this student IO branch via a KL loss; inference uses a single 5M-parameter model with one encoder. The lower strip visualizes the per-frame BIOE labels collapsing into the final segment spans.}
\label{fig:architecture}
\end{figure*}

We replace the binary boundary mask ($y_t\in\{0,1\}$) proposed by Renz et al.~\cite{renz2021segmentation} with a four-class sequence $y_t\in\{\texttt{O},\texttt{B},\texttt{I},\texttt{E}\}$ (Outside/Begin/Inside/End).
For multi-frame signs, the first and last frames are labeled \texttt{B} and \texttt{E}, with intervening frames labeled \texttt{I}; single-frame annotations (only $0.31\%/0.35\%$ of PHOENIX-14/BSL training frames) are assigned \texttt{E}.
Because the eight-transition grammar has no singleton path, their evidence contributes to the highest-scoring valid path; their rarity motivates the compact four-state representation rather than a dedicated \texttt{S} class.
The map $\texttt{O}\!\mapsto\!1$, $\{\texttt{B},\texttt{I},\texttt{E}\}\!\mapsto\!0$ recovers the binary labels exactly, so any binary boundary metric remains computable while providing explicit begin, inside, and end states for structured prediction.
At inference, we restrict decoding to eight legal frame-to-frame transitions, and a constrained Viterbi decoder~\cite{viterbi1967error} (see Sec.~\ref{sec:method-decoder}) keeps the highest emission path among them, so every decoded path satisfies the declared \bioe{} transition grammar by construction. Fig.~\ref{fig:decoder} illustrates the legal transition decoding, and Fig.~\ref{fig:architecture} summarizes the complete \bioecvr{} pipeline.

\subsection{Encoder}
Throughout our main experiments, the frame features \(X \in \mathbb{R}^{T \times d_\text{feat}}\) (\(d_\text{feat}=1024\)) are extracted using I3D~\cite{carreira2017i3d}.
The features are projected by a linear layer to $256$ dimensions, passed through a 5-block residual TCN~\cite{lea2017tcn} with dilations \([1,2,4,8,16]\) (receptive field 63 frames \(\approx 2.5\) s at 25 fps), augmented with sinusoidal positional encoding, and finally processed by a 4-layer pre-LN Transformer encoder~\cite{vaswani2017attention} (8 heads, FFN 1024).
The resulting hidden states \(H \in \mathbb{R}^{T \times 256}\) are shared by the BIOE and soft Begin/End heads and also serve as input to the distilled binary IO branch.
The IO branch adds 3 residual temporal convolution blocks with dilations $[1,2,4]$, layer normalization, and a binary sign/boundary classifier, but does not duplicate the input projection, 5-block TCN, or Transformer.
The student has 5M trainable parameters; the frozen teacher is used only during training and discarded at inference.

\subsection{Prediction Heads}
The structured decoder uses three heads on the shared states $H$.
The \textbf{fine BIOE head} $P_\text{BIOE}=\mathrm{softmax}(W_fH+b_f)\in\mathbb{R}^{T\times4}$ is the primary output (class-weighted cross-entropy, label smoothing $0.1$).
Two \textbf{soft \texttt{B}/\texttt{E} heads} $p_B,p_E=\sigma(\cdot)\in\mathbb{R}^T$ are each trained by binary cross-entropy ($\lambda_B=\lambda_E=0.4$) against Gaussian-shaped targets centered on the ground-truth begin/end frames.

\subsection{Loss Function}
Supervised training combines the 4-class loss, binary IO loss, Begin/End auxiliaries, and temporal regularization:
\[
L_\text{sup}
=
L_\text{BIOE}
+
L_\text{IO}
+
0.4(L_B+L_E)
+
0.05L_\text{smooth}.
\]
Here, \(L_\text{IO}\) is a class-weighted 2-class cross-entropy loss using the binary boundary masks of Renz et al.~\cite{renz2021segmentation} as targets. Its classes are sign/internal and boundary/outside: \(y_t^\text{IO}=1\) for \texttt{O} frames and \(0\) for \texttt{B}/\texttt{I}/\texttt{E} frames, with respective class weights \([1,4]\).
\(L_\text{smooth}\) is a \emph{boundary-aware} temporal MSE on BIOE log-probabilities,
\[
\left\|
\log P_\text{BIOE}(t)
-
\log P_\text{BIOE}(t{-}1)
\right\|_2^2,
\]
masked to zero within \(\pm2\) frames of any ground-truth boundary, penalizing interior jitter while permitting legitimate transitions.
For binary IO logits \(z_k\) define,
\[
q_k^{(T_d)}
=
\operatorname{softmax}(z_k/T_d),
\qquad
k\in\{\text{teacher},\text{student}\}.
\]
The total distillation objective~\cite{hinton2015distilling} is,
\[
L
=
L_\text{sup}
+
\lambda_\text{KD}T_d^2\,
\mathrm{KL}\!\left(
q_\text{teacher}^{(T_d)}
\Vert
q_\text{student}^{(T_d)}
\right),
\]
where \(T_d=2\) and \(\lambda_\text{KD}=1\).
Padding frames are excluded from all loss terms.

\subsection{Constrained Viterbi Decoder with B/E Fusion}
\label{sec:method-decoder}

At inference, the three structured-decoding heads are combined in steps (see Fig.~\ref{fig:decoder}):

\begin{enumerate}
\item \textbf{\texttt{B}/\texttt{E} fusion gate} (\(\alpha=0.5\), \(\tau=0.4\)):

For the \texttt{B} class, the fused probability is,
\[
\tilde{P}_B(t)=
\begin{cases}
\alpha\,p_B(t)+(1-\alpha)\,P_\text{BIOE}[B](t),
& p_B(t)\ge\tau\\
P_\text{BIOE}[B](t),
& \text{otherwise},
\end{cases}
\]
with the same operation applied to the \texttt{E} class using \(p_E(t)\) and \(P_\text{BIOE}[E](t)\).
The fused per-frame probabilities are re-normalized and log-transformed into the Viterbi emissions.
In the full \bioecvr{} model, the distilled IO branch additionally contributes its boundary prediction to the \texttt{O} class emission before decoding, as described in Sec.~\ref{sec:method-relocate}.

\item \textbf{Constrained Viterbi} finds the highest emission path through the BIOE trellis subject to eight legal transitions:
\[
    \begin{array}{ll}
    \texttt{O}\to\{\texttt{O,B}\}, & \texttt{B}\to\{\texttt{I,E}\},\\[2pt]
    \texttt{I}\to\{\texttt{I,E}\}, & \texttt{E}\to\{\texttt{O,B}\}.
    \end{array}
\]
We use a fixed \texttt{O} class emission bias \(o_{\text{bias}}=-1.0\) and transition bias \(b_{\text{trans}}=0.0\).

\item \textbf{Endpoint policy}:

The primary strict no-cut protocol uses no ground-truth endpoint cleanup. The cutting rule used by Renz et al.~\cite{renz2021segmentation} is evaluated only in a paired compatibility audit.
\end{enumerate}

\noindent The final BIOE path is converted to segments by extracting every maximal \texttt{B\,I*\,E} run.

\subsection{Distilled Endpoint Refinement}
\label{sec:method-relocate}

The constrained decoder produces structured preliminary segments, but their endpoint precision still depends on the boundary evidence available to the BIOE predictions.
To improve boundary localization while restricting changes to local endpoint positions, the student uses a binary IO branch on the shared hidden states.
For each corpus and seed, a 4.41M-parameter binary teacher is trained on the same training indices and selected by validation \mFoneB.
During student training, this teacher remains frozen and supplies softened binary targets in addition to the ground-truth binary labels.

At inference, the student's \(P(\text{boundary})\) contributes to the \texttt{O} class emission before constrained decoding using the validation-selected \(\alpha_\text{IO}\).
A boundary run is a maximal contiguous run of boundary/outside frames in the decoded binary mask.
After decoding, each run is shifted as a block toward the nearest sufficiently confident student boundary peak within the displacement radius, clipped to the video bounds, and combined with the other shifted runs; its width is preserved unless clipping or merging occurs.
For each corpus and seed, after checkpoint selection we choose
\(\alpha_\text{IO}\in\{0,0.3,0.7\}\),
peak threshold \(\tau_\text{IO}\in\{0.3,0.4,0.5\}\),
and maximum displacement \(r\in\{1,2,3\}\)
using the held-out validation score.
The selected setting is then applied once to the test set; all six selections are reported in Supplementary Tab.~\ref{tab:appx-seeds}.

This prediction-only refinement adjusts endpoint locations locally; the strict BIOE validity guarantee applies to the Viterbi path before relocation.
A 2-encoder implementation is evaluated only as a computational baseline.

\subsection{Training Protocol}
We train for 50 epochs using AdamW~\cite{loshchilov2019adamw} ($\text{lr}=3\times10^{-4}$, weight decay 0.01), with a cosine learning-rate schedule ending at $10^{-5}$, batch size 8, and gradient clipping 1.
We run three seeds (42, 43, 44) and report mean $\pm$ SD.
Student checkpoints are selected by the balanced validation score $(\mFoneB+\mFoneS)/2$ under a fixed checkpoint-selection configuration: \texttt{B}/\texttt{E} fusion $\alpha=0.5$ and $\tau=0.4$, \texttt{O} class bias $-1.0$, transition bias $0$, and IO refinement $\alpha_\text{IO}=0.7$, $\tau_\text{IO}=0.4$, and $r=3$.
After checkpoint selection, only the IO refinement parameters in Sec.~\ref{sec:method-relocate} are retuned on validation; test data are reserved for final reporting.


\section{Experimental Setup}
\label{sec:exp}

\noindent\textbf{Datasets and features.}
We use the released PHOENIX-14~\cite{koller2015phoenix14} package (4,556 train/1,115 test videos; German Sign Language) and BSL Corpus~\cite{schembri2013bsl} package (5,413/702; British Sign Language).
Both provide frame-level lexical annotations and the same 1024-D I3D~\cite{carreira2017i3d} features used by Renz et al.~\cite{renz2021segmentation}, keeping the visual representation fixed so that the controlled comparison reflects changes in the segmentation model rather than a different visual backbone.

Neither package defines a validation split for the segmentation task.
We therefore deterministically hold out $10\%$ of the original training videos using split seed 100: 4,100/456 train/validation videos for PHOENIX-14 and 4,872/541 for the BSL Corpus.
Every trained model---teachers, students, matched ablations, learned CRFs, and architecture controls---uses exactly these video indices.
The train and validation partitions are video-disjoint and together cover the original training set, although recording groups can occur in both partitions (see Sec.~\ref{sec:discussion}).

\noindent\textbf{Segmentation metrics.}
Following the evaluation protocol of Renz et al.~\cite{renz2021segmentation}, \mFoneB{} averages boundary F1 at temporal tolerances of 1--4 frames, while \mFoneS{} averages segment-IoU F1 over thresholds 0.40--0.75.
\mFoneB{} measures local boundary placement, whereas \mFoneS{} additionally penalizes fragmented, merged, or incorrectly extended segments.
All primary predictions are evaluated \emph{without} endpoint cutting.
The post-processing step for endpoint cutting used by Renz et al.~\cite{renz2021segmentation} reads ground-truth endpoints; we retain a paired cut/no-cut audit in the Supplementary only to quantify its effect, while all primary results use the no-cut protocol.

\noindent\textbf{Baselines and controls.}
The primary baseline is the checkpoint linked by the official Renz et al.~\cite{renz2021segmentation} repository, evaluated with the same strict no-cut protocol used throughout our experiments.
Because the released files do not reconstruct the paper's multi-seed aggregate, we report it as one public checkpoint rather than as the exact published mean.
Supplementary Sec.~\ref{sec:appx-cut} separately reports the published aggregates and the cut protocol compatibility audit.

On the BSL Corpus, we additionally evaluate a three-seed full-resolution reproduction of Hands-On~\cite{low2025handson} with normalized HaMeR~\cite{hamer} features at the native frame rate of 25 fps, without label cleanup or endpoint cutting.
Seed 42 uses the recovered checkpoint; seeds 43 and 44 follow the same recipe with fixed epoch-40 checkpoints.
Their predicted segments are passed to the same frozen recognizer used for the other downstream comparisons.
Because Hands-On uses HaMeR features while the controlled comparisons use I3D, it is treated as a feature-heterogeneous reference rather than a controlled I3D comparison.

Three additional architecture families use binary inside/boundary supervision: a local-attention Transformer with temporal refinement (2.51M), a BiLSTM-CRF (1.05M), and MS-TCN++ (0.89M)~\cite{li2020mstcnpp}.
They share I3D inputs, train/validation indices, seeds, 50 epochs, AdamW training setup, balanced score checkpoint selection, endpoint policy, metrics, and downstream recognizer.
Model depth and parameter count vary by architecture; the local Transformer is an ASFormer-inspired control, not an exact reproduction of its published configuration~\cite{yi2021asformer}.

\noindent\textbf{Downstream probe.}
Each predicted segment sequence is passed to one frozen recognizer per corpus: the released PHOENIX-14 classifier~\cite{renz2021segmentation} and a 979-class BSL classifier shared across all BSL experiments.
The classifier receives the frames selected by a segmenter, while its weights, vocabulary, and decoding remain unchanged.
We report corpus \wer$=(S+D+I)/N$ and its substitution, deletion, and insertion components, with fixed denominators $N$: 10,987 (PHOENIX-14) and 4,214 (BSL Corpus).
Test \wer and recognition accuracy are never used for checkpoint or decoder selection.

\noindent\textbf{Statistics.}
New models report three-seed mean and sample standard deviation.
The BIO/BIOE comparison is paired by seed and emphasizes paired differences because three pairs cannot support a high-powered significance test.
For the five family metric analysis, the unit is one architecture-family mean; with $n=5$, we use exact two-sided permutation $p$-values for Spearman $\rho$ and report the small-sample limitation explicitly.

\section{Results}
\label{sec:results}

\subsection{No-Cut Segmentation and Recognition}

\begin{table}[t]
\centering
\small
\setlength{\tabcolsep}{3.0pt}
\caption{Strict no-cut segmentation results. \bioecvr{} and Hands-On~\cite{low2025handson}$^{\dagger}$ are three-seed mean $\pm$ SD; Renz et al.~\cite{renz2021segmentation} is the public checkpoint. Hands-On$^{\dagger}$ uses full-resolution HaMeR features; all other rows use I3D.}
\label{tab:main}
\begin{tabular}{@{}llcc@{}}
\toprule
\textbf{Corpus} & \textbf{Segmenter} & \textbf{mF1B$\uparrow$} & \textbf{mF1S$\uparrow$} \\
\midrule
\multirow{2}{*}{PHX~\cite{koller2015phoenix14}}
& Renz et al.~\cite{renz2021segmentation} & 69.76 & 51.12 \\
& \bioecvr{} & \textbf{71.93$\pm$0.19} & \textbf{65.70$\pm$0.15} \\
\midrule
\multirow{3}{*}{BSL~\cite{schembri2013bsl}}
& Renz et al.~\cite{renz2021segmentation} & 67.28 & 43.83 \\
& Hands-On~\cite{low2025handson}$^{\dagger}$ & 62.39$\pm$0.06 & 47.06$\pm$0.41 \\
& \bioecvr{} & \textbf{68.50$\pm$0.64} & \textbf{58.42$\pm$1.03} \\
\bottomrule
\end{tabular}
\end{table}

Under the same strict no-cut endpoint protocol, \bioecvr{} improves over the public Renz et al.~\cite{renz2021segmentation} checkpoint by $\mathord{+}2.17/\mathord{+}14.58$ \mFoneB/\mFoneS{} on PHOENIX-14 and $\mathord{+}1.22/\mathord{+}14.59$ on the BSL Corpus (see Tab.~\ref{tab:main}).
On the BSL Corpus, it also exceeds our three-seed full-resolution reproduction of Hands-On~\cite{low2025handson} by $\mathord{+}6.11/\mathord{+}11.36$ \mFoneB/\mFoneS{}; because Hands-On uses HaMeR while the other rows use I3D, this is a feature-heterogeneous reference rather than a controlled I3D comparison.

In downstream recognition, \bioecvr{} reduces \wer relative to the public Renz et al.~\cite{renz2021segmentation} checkpoint by 9.10 points ($18.3\%$ relative) on PHOENIX-14 and 4.15 points ($5.36\%$ relative) on the BSL Corpus.
Under the same frozen recognizer, the Hands-On~\cite{low2025handson} reproduction yields $77.52\pm0.41$ \wer, whereas \bioecvr{} achieves $73.33\pm0.31$, a 4.19-point ($5.40\%$ relative) reduction in mean \wer (see Tab.~\ref{tab:wer}).

\begin{table*}[!t]
\centering
\small
\setlength{\tabcolsep}{5.2pt}
\caption{Downstream recognition under a matched recognizer. \wer and substitution/deletion/insertion rates are corpus-level percentages; lower is better. \bioecvr{} and Hands-On$^{\dagger}$ use three-seed mean $\pm$ SD for \wer and three-seed mean for S/D/I. Hands-On$^{\dagger}$ uses HaMeR features; dashes denote an unavailable PHOENIX-14 model. None uses test labels for model/decoder selection or endpoint cutting.}
\label{tab:wer}
\begin{tabular}{lcccccccc}
\toprule
& \multicolumn{4}{c}{\textbf{PHX~\cite{koller2015phoenix14}}} & \multicolumn{4}{c}{\textbf{BSL~\cite{schembri2013bsl}}} \\
\cmidrule(lr){2-5}\cmidrule(lr){6-9}
\textbf{Segmenter} & \textbf{WER$\downarrow$} & \textbf{S} & \textbf{D} & \textbf{I} & \textbf{WER$\downarrow$} & \textbf{S} & \textbf{D} & \textbf{I} \\
\midrule
Renz et al.~\cite{renz2021segmentation} & 49.73 & 32.03 & 5.55 & 12.15 & 77.48 & 57.85 & 12.13 & 7.50 \\
Hands-On~\cite{low2025handson}$^{\dagger}$ & --- & --- & --- & --- & 77.52$\pm$0.41 & 55.55 & 16.55 & 5.42 \\
\bioecvr{} & \textbf{40.63$\pm$0.20} & 27.06 & 8.90 & 4.67 & \textbf{73.33$\pm$0.31} & 55.43 & 12.68 & 5.22 \\
\bottomrule
\end{tabular}
\end{table*}

Across three seeds, on PHOENIX-14 \mFoneS/\wer spans 65.54--65.83/40.41--40.79, and on the BSL Corpus spans 57.39--59.45/73.02--73.64.
On PHOENIX-14, insertion/substitution errors fall by 7.48/4.97 points relative to Renz et al.~\cite{renz2021segmentation} while deletion rises by 3.35, consistent with fewer fragmented predictions and occasional missed short signs.
On the BSL Corpus, insertion/substitution fall by 2.28/2.42 points while deletion rises by 0.55.

\begin{figure}[t]
\centering
\includegraphics[width=\columnwidth]{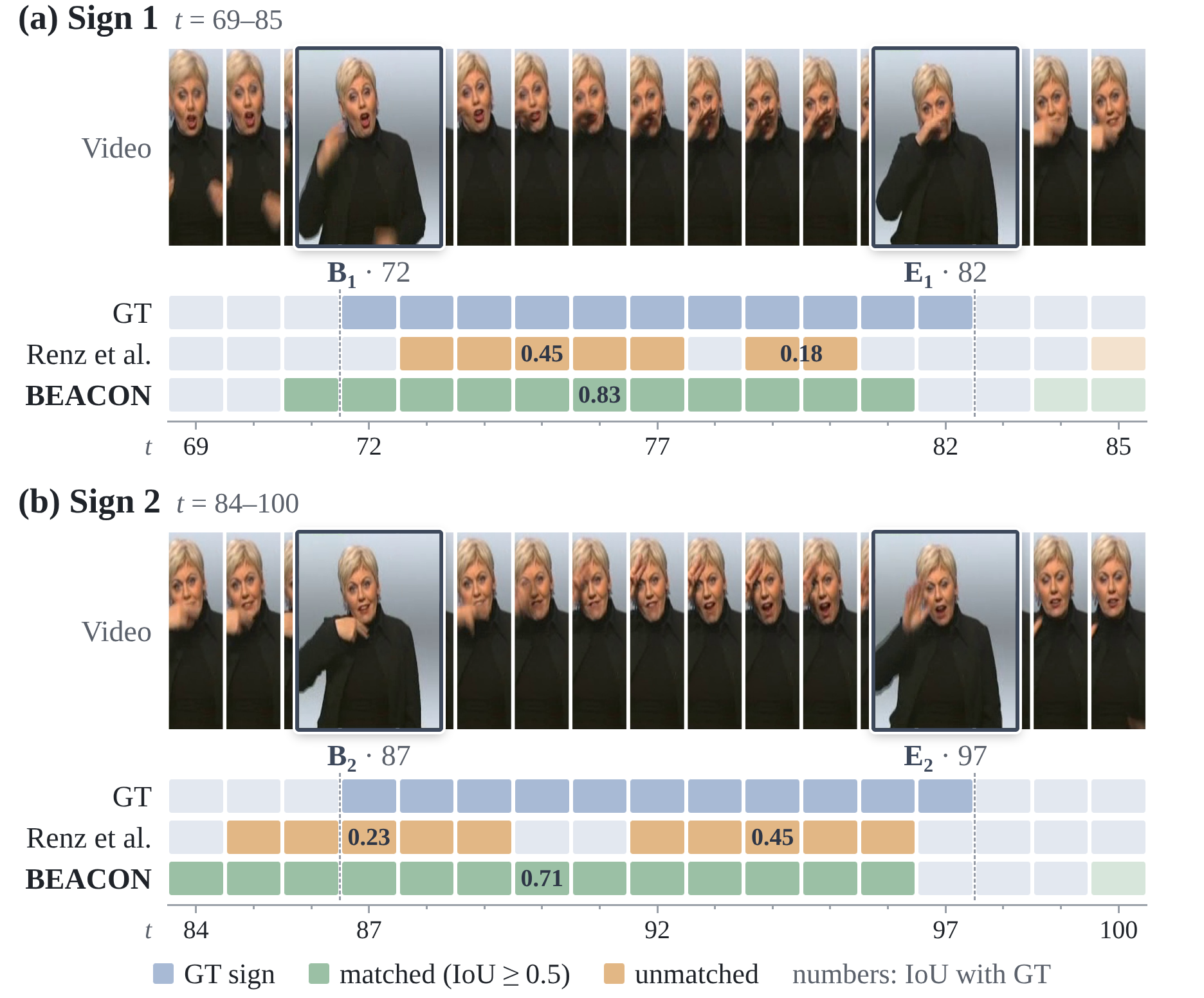}
\caption{PHOENIX-14~\cite{koller2015phoenix14} example (no-cut; seed 42). Enlarged B/E frames anchor each clip to the frame-aligned GT and predictions. Numbers report segment IoU with the corresponding GT sign.}
\label{fig:qual}
\end{figure}

Figure~\ref{fig:qual} gives a frame-aligned view of two neighboring ground-truth signs. Renz et al.~\cite{renz2021segmentation} divide each sign into two unmatched fragments (IoUs 0.45/0.18 and 0.23/0.45), whereas \bioecvr{} produces one matched segment per sign (IoUs 0.83/0.71).

Endpoint cutting has little effect on \bioecvr{}: relative to no-cut, the paired cut evaluation changes \wer by $\mathord{-}0.05/\mathord{-}0.16$ and \mFoneS{} by $\mathord{+}0.08/\mathord{+}0.24$ on the PHOENIX-14/BSL Corpus.
This indicates that the primary results do not depend on ground-truth endpoint cleanup.
For the public Renz et al.~\cite{renz2021segmentation} checkpoint, cutting increases \mFoneS{} by $\mathord{+}0.54/\mathord{+}1.17$, showing that the cut and no-cut protocols are not equivalent for this baseline.
Per-seed results and the full paired compatibility audit are provided in the Supplementary.

\subsection{Architecture Families and Metric Associations}
\label{sec:results-families}
The association analysis (see Tab.~\ref{tab:families}) uses one representative point from each of five distinct architecture families (see Supplementary Tab.~\ref{tab:appx-families}). All newly trained controls share features, split, training budget, selection rule, no-cut evaluation, and frozen recognizer.

Across these architectures, \mFoneS{} has a stronger negative Spearman association with \wer: $\rho=-1.0$ versus $-0.9$ on PHOENIX-14 and $-0.9$ versus $-0.8$ on BSL. The exact test reaches $p<0.05$ only for PHOENIX-14 \mFoneS; hence, the evidence supports only that \mFoneS{} orders \wer more consistently among the tested families. Leave-one-family-out \mFoneS{} results are $\rho=-1.0$ throughout on PHOENIX-14 and $[-1.00,-0.80]$ on BSL (see Supplementary). On BSL, MS-TCN++ has 0.21 lower mean \wer than \bioecvr{} despite 2.64 lower \mFoneS, within either method's reported seed variation.

\begin{table}[t]
\centering
\small
\setlength{\tabcolsep}{3.4pt}
\caption{Architecture families comparison under strict no-cut evaluation (three-seed mean $\pm$ SD except Renz et al.~\cite{renz2021segmentation}'s MS-TCN). Local Transformer is an architecture control, not an exact ASFormer~\cite{yi2021asformer} reproduction.}
\label{tab:families}
\begin{tabular}{lccc}
\toprule
\multicolumn{4}{c}{\textbf{PHX~\cite{koller2015phoenix14}}} \\
\cmidrule(lr){1-4}
\textbf{Family} & \textbf{mF1B$\uparrow$} & \textbf{mF1S$\uparrow$} & \textbf{WER$\downarrow$} \\
\midrule
MS-TCN & 69.76 & 51.12 & 49.73 \\
Local Transformer & 48.29$\pm$0.54 & 37.03$\pm$0.07 & 50.90$\pm$0.92 \\
BiLSTM-CRF & 67.80$\pm$0.10 & 58.08$\pm$0.26 & 47.15$\pm$0.75 \\
MS-TCN++ & 70.60$\pm$0.67 & 60.10$\pm$1.71 & 42.54$\pm$0.29 \\
\bioecvr{} & \textbf{71.93$\pm$0.19} & \textbf{65.70$\pm$0.15} & \textbf{40.63$\pm$0.20} \\
\midrule
\multicolumn{4}{c}{\textbf{BSL~\cite{schembri2013bsl}}} \\
\cmidrule(lr){1-4}
\textbf{Family} & \textbf{mF1B$\uparrow$} & \textbf{mF1S$\uparrow$} & \textbf{WER$\downarrow$} \\
\midrule
MS-TCN & 67.28 & 43.83 & 77.48 \\
Local Transformer & 49.97$\pm$1.16 & 36.36$\pm$0.84 & 79.41$\pm$0.49 \\
BiLSTM-CRF & 62.59$\pm$0.26 & 50.94$\pm$0.15 & 76.15$\pm$0.13 \\
MS-TCN++ & 67.68$\pm$1.33 & 55.78$\pm$0.93 & \textbf{73.12$\pm$0.63} \\
\bioecvr{} & \textbf{68.50$\pm$0.64} & \textbf{58.42$\pm$1.03} & 73.33$\pm$0.31 \\
\bottomrule
\end{tabular}
\end{table}

\subsection{Controlled Analyses}
\label{sec:results-ablations}
\noindent\textbf{Matched BIO/BIOE.}
To isolate labels and decoding, we hold backbone, optimizer, split, loss, seeds, and 50-epoch budget fixed and cross BIO/BIOE labeling with argmax or the corresponding grammar-constrained decoder (see Tab.~\ref{tab:ablation}). Under argmax, changing BIO to BIOE changes \mFoneS{} by $-0.01/+0.58$ on PHOENIX-14/BSL; with each representation's matching grammar, the changes are $-0.04/+0.92$, and \wer changes by $-0.53/0.00$. The isolated effect is modest and dataset-dependent. BIOE nevertheless represents a complete segment directly as a \texttt{B\,I*\,E} run. The full \bioecvr{} improvement therefore reflects the combined system rather than the label alphabet alone.

Relative to argmax, matching-grammar decoding changes \mFoneS{} by $+0.51/+0.02$ for BIOE and $+0.54/-0.31$ for BIO. Its numerical effect is also modest and dataset-dependent; structurally, however, argmax can leave the declared grammar whereas constrained decoding cannot.

Relative to BIO/argmax in Tab.~\ref{tab:ablation}, the full model gains $+1.15/+3.71$ \mFoneB{} and $+3.44/+3.57$ \mFoneS{} on PHOENIX-14/BSL. Tab.~\ref{tab:distill-diagnostic} separates endpoint modeling from distillation. The supervised IO branch and refinement add $+1.17/+2.62$ \mFoneB{} and $+2.46/+2.72$ \mFoneS; KD then adds $+0.38/+0.40$ and $+0.48/+0.25$, respectively. The total gains over the structured base are $+1.55/+3.02$ \mFoneB{} and $+2.94/+2.97$ \mFoneS; the table also reports the corresponding \wer.

\begin{table}[t]
\centering
\scriptsize
\setlength{\tabcolsep}{2.0pt}
\caption{Matched label/readout comparison, strict no-cut (three-seed mean). Other factors are fixed; each constrained row uses the matching grammar. \wer is reported for constrained readout.}
\label{tab:ablation}
\begin{tabular}{llrrrr}
\toprule
\textbf{Corpus} & \textbf{Labels/Readout} & \textbf{mF1B$\uparrow$} & \textbf{mF1S$\uparrow$} & $\Delta$\textbf{mF1S} & \textbf{WER$\downarrow$} \\
\midrule
\multirow{4}{*}{PHX~\cite{koller2015phoenix14}} & BIO/argmax & 70.78 & 62.26 & --- & --- \\
& BIO/grammar & 70.68 & 62.80 & --- & 43.82 \\
& BIOE/argmax & 70.56 & 62.25 & --- & --- \\
& BIOE/8-edge & 70.38 & 62.76 & $-0.04$ & 43.29 \\
\midrule
\multirow{4}{*}{BSL~\cite{schembri2013bsl}} & BIO/argmax & 64.79 & 54.85 & --- & --- \\
& BIO/grammar & 63.98 & 54.54 & --- & 73.95 \\
& BIOE/argmax & 66.27 & 55.43 & --- & --- \\
& BIOE/8-edge & 65.48 & 55.45 & $+0.92$ & 73.95 \\
\bottomrule
\end{tabular}
\end{table}

\begin{table}[t]
\centering
\scriptsize
\setlength{\tabcolsep}{3.0pt}
\caption{Component comparison, strict no-cut (three-seed mean). Each entry reports PHOENIX-14/BSL. The last two rows have the same shared-student architecture and endpoint readout, isolating KD.}
\label{tab:distill-diagnostic}
\begin{tabular}{lrrr}
\toprule
\textbf{System} & \textbf{mF1B$\uparrow$} & \textbf{mF1S$\uparrow$} & \textbf{WER$\downarrow$} \\
\midrule
BIOE/8-edge base & 70.38/65.48 & 62.76/55.45 & 43.29/73.95 \\
IO branch, no KD & 71.55/68.10 & 65.22/58.17 & 40.72/73.74 \\
Full \bioecvr{} & \textbf{71.93/68.50} & \textbf{65.70/58.42} & \textbf{40.63/73.33} \\
\bottomrule
\end{tabular}
\end{table}

\noindent\textbf{Fixed grammar and learned transitions.}
A same-backbone unrestricted CRF is numerically competitive (see Supplementary Tab.~\ref{tab:appx-crf}). It is close to fixed decoding on PHOENIX-14 and improves the matched non-distilled BIOE control on the BSL Corpus. However, 12.65\%/12.63\% of its decoded test sequences contain at least one transition outside the fixed grammar, whereas constrained decoding guarantees zero grammar-invalid sequences. Learned transitions are viable and can be numerically competitive; the fixed graph instead provides deterministic validity under the declared BIOE grammar.

\noindent\textbf{Inference cost.}
Supplementary Tab.~\ref{tab:appx-efficiency} compares a two-encoder implementation with the shared student. At the p50/p90 sequence lengths of both corpora, the student reduces neural-forward latency by $36.5\%$--$36.9\%$ and peak allocation by roughly $29\%$--$33\%$. The teacher is never loaded at inference. Absolute timings are hardware-dependent, making the relative reductions more informative.

\section{Discussion}
\label{sec:discussion}

\subsection{Segment Quality and Downstream Recognition}
\label{sec:discussion-mf1b}

\bioecvr{} improves boundary and segment metrics under strict no-cut evaluation, with the larger gain at the segment level, and the same predictions lower \wer under the frozen recognizers. The paired audit confirms that this pattern does not depend on ground-truth endpoint cleanup.

On PHOENIX-14, fewer insertions are consistent with fewer fragmented or spurious segments, while more deletions indicate occasional missed signs. On BSL, gains span substitutions and insertions. Together with Sec.~\ref{sec:results-families}, this motivates reporting \mFoneB, \mFoneS, and downstream \wer jointly.

\subsection{Structure and Boundary Synergy}
\label{sec:discussion-bioe-io}

Sec.~\ref{sec:results-ablations} shows modest standalone numerical effects for BIOE and constrained decoding. BIOE directly represents completion as a \texttt{B\,I*\,E} run; constrained decoding enforces the declared grammar. Learned CRF transitions can be competitive, whereas the fixed graph guarantees validity under that grammar.

Structural validity does not determine exact endpoints. The supervised IO branch and refinement provide most of the gain beyond the structured base; distillation adds a smaller, consistent gain without changing inference. \bioecvr{} thereby combines valid structure and precise endpoint localization in one model.

\subsection{Limitations}

\noindent\textbf{Validation and generalization.}
Neither dataset defines an official segmentation validation split, so we construct a fixed video-disjoint partition from its training pool. It is not recording-group-disjoint; generalization to unseen recording groups remains untested.

\noindent\textbf{Scope of the metric--WER evidence.}
The five-family analysis uses one frozen recognizer per corpus. This small sample is corpus- and recognizer-specific and cannot establish universal recognition optimality.

\noindent\textbf{Dense supervision and training complexity.}
\bioecvr{} requires frame-level annotations and does not address corpora with only sentence- or gloss-level supervision. Inference uses one student, but training also requires a binary teacher and validation-selected refinement parameters.

\noindent\textbf{Annotation scope.}
Frame endpoints are operational annotations, not unique linguistic boundaries: co-articulation and transitions may be ambiguous, and the corpora provide no inter-annotator analysis.

\section{Conclusion}
\label{sec:conclusion}

We presented \bioecvr, a single-encoder segmenter that combines four-class \bioe{} labeling, eight-transition constrained decoding, and distilled boundary knowledge. The design separates two complementary requirements: the fixed grammar restricts predictions to valid BIOE paths, while the shared IO branch supplies local evidence for endpoint refinement. Its binary teacher is used only during training, so inference retains one 5M-parameter student rather than a separate boundary encoder.

Under strict no-cut evaluation with validation-based selection, \bioecvr{} improves \mFoneS{} by $14.58/14.59$ points and reduces downstream \wer by $9.10/4.15$ points relative to the state-of-the-art on PHOENIX-14/BSL. The matched ablations show that the full gain is a system-level result: the shared endpoint branch provides the largest improvement beyond the structured base, and distillation adds a smaller consistent gain without increasing inference cost. Across five architecture families, \mFoneS{} orders \wer more consistently than \mFoneB. Although this evidence is limited to the tested corpora and recognizers, it supports reporting segment quality, boundary accuracy, and downstream \wer together rather than selecting models by boundary precision alone.

Beyond sign segmentation, fixed-grammar decoding and training-only boundary distillation may be useful when temporal outputs must satisfy a known structure but local endpoints remain uncertain. Future work should model annotation ambiguity and transition regions, reduce dependence on dense frame-level supervision, broaden matched evaluation across datasets, recognizers, and vocabularies, and train segmenters more directly for downstream utility.

\FloatBarrier

\clearpage

\iftoggle{wacvfinal}{\balance}{}
{\footnotesize
\bibliographystyle{ieeenat_fullname}
\bibliography{references}

\begin{thebibliography}{26}
\providecommand{\natexlab}[1]{#1}
\providecommand{\url}[1]{\texttt{#1}}
\expandafter\ifx\csname urlstyle\endcsname\relax
  \providecommand{\doi}[1]{doi: #1}\else
  \providecommand{\doi}{doi: \begingroup \urlstyle{rm}\Url}\fi

\bibitem[Abu~Farha and Gall(2019)]{farha2019mstcn}
Yazan Abu~Farha and Juergen Gall.
\newblock {MS-TCN}: Multi-stage temporal convolutional network for action
  segmentation.
\newblock In \emph{IEEE Conference on Computer Vision and Pattern Recognition
  (CVPR)}, pages 3575--3584, 2019.

\bibitem[Belissen et~al.(2020)Belissen, Braffort, and
  Gouiff{\`e}s]{belissen2020dicta}
Valentin Belissen, Annelies Braffort, and Mich{\`e}le Gouiff{\`e}s.
\newblock {Dicta-Sign-LSF-v2}: Remake of a continuous {F}rench {S}ign
  {L}anguage dialogue corpus and a first baseline for automatic sign language
  processing.
\newblock In \emph{Proceedings of the Twelfth Language Resources and Evaluation
  Conference (LREC)}, pages 6040--6048. European Language Resources
  Association, 2020.

\bibitem[Carreira and Zisserman(2017)]{carreira2017i3d}
Joao Carreira and Andrew Zisserman.
\newblock Quo vadis, action recognition? {A} new model and the {K}inetics
  dataset.
\newblock In \emph{IEEE Conference on Computer Vision and Pattern Recognition
  (CVPR)}, pages 6299--6308, 2017.

\bibitem[Farfan et~al.(2025)Farfan, Choquehuanca-Zevallos, Aguilera, Dongo, and
  Cardinale]{farfan2025peru}
Summy Farfan, Juan~J. Choquehuanca-Zevallos, Ana Aguilera, Irvin Dongo, and
  Yudith Cardinale.
\newblock Temporal video segmentation approach for peruvian sign language.
\newblock \emph{Sensors}, 25\penalty0 (17):\penalty0 5217, 2025.

\bibitem[Graves et~al.(2006)Graves, Fern\'andez, Gomez, and
  Schmidhuber]{graves2006ctc}
Alex Graves, Santiago Fern\'andez, Faustino Gomez, and J\"urgen Schmidhuber.
\newblock Connectionist temporal classification: Labelling unsegmented sequence
  data with recurrent neural networks.
\newblock In \emph{International Conference on Machine Learning (ICML)}, 2006.

\bibitem[Hinton et~al.(2015)Hinton, Vinyals, and Dean]{hinton2015distilling}
Geoffrey Hinton, Oriol Vinyals, and Jeff Dean.
\newblock Distilling the knowledge in a neural network, 2015.

\bibitem[Koller et~al.(2015)Koller, Forster, and Ney]{koller2015phoenix14}
Oscar Koller, Jens Forster, and Hermann Ney.
\newblock Continuous sign language recognition: Towards large vocabulary
  statistical recognition systems handling multiple signers.
\newblock \emph{Computer Vision and Image Understanding}, 141:\penalty0
  108--125, 2015.

\bibitem[Lafferty et~al.(2001)Lafferty, McCallum, and Pereira]{lafferty2001crf}
John~D. Lafferty, Andrew McCallum, and Fernando C.~N. Pereira.
\newblock Conditional random fields: Probabilistic models for segmenting and
  labeling sequence data.
\newblock In \emph{International Conference on Machine Learning (ICML)}, pages
  282--289, 2001.

\bibitem[Lample et~al.(2016)Lample, Ballesteros, Subramanian, Kawakami, and
  Dyer]{lample2016neural}
Guillaume Lample, Miguel Ballesteros, Sandeep Subramanian, Kazuya Kawakami, and
  Chris Dyer.
\newblock Neural architectures for named entity recognition.
\newblock In \emph{Proceedings of the 2016 Conference of the North American
  Chapter of the Association for Computational Linguistics: Human Language
  Technologies (NAACL-HLT)}, pages 260--270, 2016.

\bibitem[Lea et~al.(2017)Lea, Flynn, Vidal, Reiter, and Hager]{lea2017tcn}
Colin Lea, Michael~D. Flynn, Rene Vidal, Austin Reiter, and Gregory~D. Hager.
\newblock Temporal convolutional networks for action segmentation and
  detection.
\newblock In \emph{IEEE Conference on Computer Vision and Pattern Recognition
  (CVPR)}, pages 1003--1012, 2017.

\bibitem[Li et~al.(2023)Li, Abu~Farha, Liu, Cheng, and Gall]{li2020mstcnpp}
Shijie Li, Yazan Abu~Farha, Yun Liu, Ming-Ming Cheng, and Juergen Gall.
\newblock {MS-TCN++}: Multi-stage temporal convolutional network for action
  segmentation.
\newblock \emph{IEEE Transactions on Pattern Analysis and Machine
  Intelligence}, 45\penalty0 (6):\penalty0 6647--6658, 2023.

\bibitem[Loshchilov and Hutter(2019)]{loshchilov2019adamw}
Ilya Loshchilov and Frank Hutter.
\newblock Decoupled weight decay regularization.
\newblock In \emph{International Conference on Learning Representations
  (ICLR)}, 2019.

\bibitem[Low et~al.(2025)Low, Walsh, Sincan, and Bowden]{low2025handson}
Jian~He Low, Harry Walsh, Ozge~Mercanoglu Sincan, and Richard Bowden.
\newblock {Hands-On}: Segmenting individual signs from continuous sequences.
\newblock In \emph{IEEE International Conference on Automatic Face and Gesture
  Recognition (FG)}, pages 1--5, 2025.

\bibitem[Moryossef et~al.(2023)Moryossef, Jiang, M\"uller, Ebling, and
  Goldberg]{moryossef2023seg}
Amit Moryossef, Zifan Jiang, Mathias M\"uller, Sarah Ebling, and Yoav Goldberg.
\newblock Linguistically motivated sign language segmentation.
\newblock In \emph{Findings of the Association for Computational Linguistics:
  EMNLP 2023}, pages 12703--12724. Association for Computational Linguistics,
  2023.

\bibitem[Neidle and Metaxas(2025)]{asllrp}
Carol Neidle and Dimitris Metaxas.
\newblock {American Sign Language Linguistic Research Project (ASLLRP) Sign
  Bank: ASLLRP Continuous Signing Corpora --- version 3}, 2025.
\newblock Videos provided by DawnSignPress (\url{www.dawnsign.com}).

\bibitem[Pavlakos et~al.(2024)Pavlakos, Shan, Radosavovic, Kanazawa, Fouhey,
  and Malik]{hamer}
Georgios Pavlakos, Dandan Shan, Ilija Radosavovic, Angjoo Kanazawa, David
  Fouhey, and Jitendra Malik.
\newblock Reconstructing hands in {3D} with transformers.
\newblock In \emph{IEEE/CVF Conference on Computer Vision and Pattern
  Recognition (CVPR)}, 2024.

\bibitem[Ramshaw and Marcus(1995)]{ramshaw1995text}
Lance~A. Ramshaw and Mitchell~P. Marcus.
\newblock Text chunking using transformation-based learning.
\newblock In \emph{Third Workshop on Very Large Corpora}, pages 82--94, 1995.

\bibitem[Ratinov and Roth(2009)]{ratinov2009design}
Lev Ratinov and Dan Roth.
\newblock Design challenges and misconceptions in named entity recognition.
\newblock In \emph{Proceedings of the Thirteenth Conference on Computational
  Natural Language Learning (CoNLL)}, pages 147--155, 2009.

\bibitem[Renz et~al.(2021{\natexlab{a}})Renz, Stache, Albanie, and
  Varol]{renz2021segmentation}
Katrin Renz, Nicolaj~C. Stache, Samuel Albanie, and G{\"u}l Varol.
\newblock Sign language segmentation with temporal convolutional networks.
\newblock In \emph{IEEE International Conference on Acoustics, Speech and
  Signal Processing (ICASSP)}, pages 2135--2139, 2021{\natexlab{a}}.

\bibitem[Renz et~al.(2021{\natexlab{b}})Renz, Stache, Fox, Varol, and
  Albanie]{renz2021cmpl}
Katrin Renz, Nicolaj~C. Stache, Neil Fox, G\"ul Varol, and Samuel Albanie.
\newblock Sign segmentation with changepoint-modulated pseudo-labelling.
\newblock In \emph{IEEE/CVF Conference on Computer Vision and Pattern
  Recognition Workshops (CVPRW)}, pages 3403--3412, 2021{\natexlab{b}}.

\bibitem[Schembri et~al.(2013)Schembri, Fenlon, Rentelis, Reynolds, and
  Cormier]{schembri2013bsl}
Adam Schembri, Jordan Fenlon, Ramas Rentelis, Sally Reynolds, and Kearsy
  Cormier.
\newblock Building the {B}ritish {S}ign {L}anguage corpus.
\newblock \emph{Language Documentation \& Conservation}, 7:\penalty0 136--154,
  2013.

\bibitem[Sim et~al.(2024)Sim, Cho, and Lee]{sim2024changepoint}
HoHyun Sim, Hyeonjoong Cho, and Hankyu Lee.
\newblock Changepoint detection-assisted nonparametric clustering for
  unsupervised temporal sign segmentation.
\newblock \emph{Engineering Applications of Artificial Intelligence},
  127:\penalty0 107323, 2024.

\bibitem[Vaswani et~al.(2017)Vaswani, Shazeer, Parmar, Uszkoreit, Jones, Gomez,
  Kaiser, and Polosukhin]{vaswani2017attention}
Ashish Vaswani, Noam Shazeer, Niki Parmar, Jakob Uszkoreit, Llion Jones,
  Aidan~N. Gomez, Lukasz Kaiser, and Illia Polosukhin.
\newblock Attention is all you need.
\newblock In \emph{Advances in Neural Information Processing Systems
  (NeurIPS)}, pages 5998--6008, 2017.

\bibitem[Viterbi(1967)]{viterbi1967error}
Andrew~J. Viterbi.
\newblock Error bounds for convolutional codes and an asymptotically optimum
  decoding algorithm.
\newblock \emph{IEEE Transactions on Information Theory}, 13\penalty0
  (2):\penalty0 260--269, 1967.

\bibitem[Yi et~al.(2021)Yi, Wen, and Jiang]{yi2021asformer}
Fangqiu Yi, Hongyu Wen, and Tingting Jiang.
\newblock {ASFormer}: Transformer for action segmentation.
\newblock In \emph{British Machine Vision Conference (BMVC)}, 2021.

\bibitem[Zhao et~al.(2025)Zhao, Yang, Zhou, Xia, Jin, He, Neidle, and
  Metaxas]{zhao2025mhb}
Mingyu Zhao, Zhanfu Yang, Yang Zhou, Zhaoyang Xia, Can Jin, Xiaoxiao He, Carol
  Neidle, and Dimitris~N. Metaxas.
\newblock {MHB}: Multimodal handshape-aware boundary detection for continuous
  sign language recognition.
\newblock \emph{arXiv preprint arXiv:2511.19907}, 2025.

\end{thebibliography}
}

\clearpage
\appendix
\section{Additional Experimental Details}

\subsection{Canonical No-Cut Results by Seed}
\label{sec:appx-perseed}
Tab.~\ref{tab:appx-seeds} gives every run behind the primary results.
For each dataset, the binary teacher and distilled student use the same fixed 90/10 split of the original training set (split seed 100).
The teacher and student checkpoints and IO refinement parameters are selected on validation segmentation metrics; the \texttt{B}/\texttt{E} fusion gate is fixed across runs.
Seed-43 on PHOENIX-14 selects $\alpha_{\rm IO}=0$, yet still retains the IO/KD training losses.
This suggests that distillation can regularize the shared encoder even when validation does not select IO fusion at inference.

\begin{table}[H]
\centering
\scriptsize
\setlength{\tabcolsep}{3pt}
\caption{Distilled \bioecvr per-seed segmentation results and validation-selected IO refinement settings $(\alpha_{\rm IO},r,\tau_{\rm IO})$, all under strict no-cut evaluation.}
\label{tab:appx-seeds}
\begin{tabular}{lrrrrrr}
\toprule
\textbf{Corpus/Seed} & \textbf{Epoch} & $\boldsymbol{\alpha}$ & $\boldsymbol{r}$ & $\boldsymbol{\tau}$ & \textbf{mF1B$\uparrow$} & \textbf{mF1S$\uparrow$} \\
\midrule
PHX/42 & 12 & .7 & 2 & .5 & 71.80 & 65.54 \\
PHX/43 & 6  & .0 & 1 & .3 & 71.85 & 65.83 \\
PHX/44 & 10 & .7 & 3 & .3 & 72.15 & 65.74 \\
\textbf{mean$\pm$SD} & --- & --- & --- & --- & \textbf{71.93$\pm$.19} & \textbf{65.70$\pm$.15} \\
\midrule
BSL/42 & 10 & .7 & 3 & .5 & 67.92 & 57.39 \\
BSL/43 & 6  & .7 & 1 & .3 & 69.18 & 58.42 \\
BSL/44 & 3  & .7 & 3 & .5 & 68.41 & 59.45 \\
\textbf{mean$\pm$SD} & --- & --- & --- & --- & \textbf{68.50$\pm$.64} & \textbf{58.42$\pm$1.03} \\
\bottomrule
\end{tabular}
\end{table}

\begin{table}[H]
\centering
\scriptsize
\caption{Hands-On$^{\dagger}$ reproduction by seed using full-resolution HaMeR features under strict no-cut evaluation. Seed-42 uses the recovered checkpoint; seeds-43 and 44 use fixed epoch-40 checkpoints from the same recipe.}
\label{tab:appx-handson-seg}
\begin{tabular}{lrr}
\toprule
\textbf{Seed} & \textbf{mF1B$\uparrow$} & \textbf{mF1S$\uparrow$} \\
\midrule
42 & 62.45 & 47.44 \\
43 & 62.38 & 47.11 \\
44 & 62.33 & 46.63 \\
\textbf{mean$\pm$SD} & \textbf{62.39$\pm$.06} & \textbf{47.06$\pm$.41} \\
\bottomrule
\end{tabular}
\end{table}

\subsection{Per-Run Downstream Recognition}
\label{sec:appx-downstream}
Tab.~\ref{tab:appx-downstream} separates the downstream error decomposition from the segmentation and decoder selection results above.
Hands-On$^{\dagger}$ uses seeds 42--44 at full 25 fps resolution with checkpoint-matched normalized HaMeR features, raw BIO argmax, no label cleanup, and no endpoint cutting. Seed-42 is the recovered checkpoint; seeds 43 and 44 follow the same recipe and use fixed epoch-40 checkpoints. Their segments index the same I3D stream consumed by the frozen recognizer. All BSL rows use $N=4{,}214$ and all PHX rows use $N=10{,}987$.

\begin{table}[H]
\centering
\scriptsize
\setlength{\tabcolsep}{3.5pt}
\caption{Per-run downstream decomposition under the same frozen recognizer per corpus. Rates are corpus-level percentages. Renz et al. is one public checkpoint; Hands-On$^{\dagger}$ and \bioecvr report all three seeds, with mean $\pm$ SD for WER and mean S/D/I.}
\label{tab:appx-downstream}
\begin{tabular}{llrrrr}
\toprule
\textbf{Corpus} & \textbf{Segmenter/Run} & \textbf{WER} & \textbf{S} & \textbf{D} & \textbf{I} \\
\midrule
\multirow{5}{*}{PHX}
& Renz et al. & 49.73 & 32.03 & 5.55 & 12.15 \\
& BEACON/42 & 40.79 & 27.21 & 8.87 & 4.71 \\
& BEACON/43 & 40.41 & 26.60 & 10.13 & 3.68 \\
& BEACON/44 & 40.69 & 27.37 & 7.71 & 5.62 \\
& \textbf{BEACON mean$\pm$SD} & \textbf{40.63$\pm$.20} & 27.06 & 8.90 & 4.67 \\
\midrule
\multirow{9}{*}{BSL}
& Renz et al. & 77.48 & 57.85 & 12.13 & 7.50 \\
& Hands-On$^{\dagger}$/42 & 77.98 & 55.93 & 15.95 & 6.10 \\
& Hands-On$^{\dagger}$/43 & 77.41 & 55.67 & 16.94 & 4.79 \\
& Hands-On$^{\dagger}$/44 & 77.17 & 55.05 & 16.75 & 5.36 \\
& \textbf{Hands-On$^{\dagger}$ mean$\pm$SD} & 77.52$\pm$.41 & 55.55 & 16.55 & 5.42 \\
& BEACON/42 & 73.33 & 55.08 & 13.50 & 4.75 \\
& BEACON/43 & 73.64 & 55.65 & 11.96 & 6.03 \\
& BEACON/44 & 73.02 & 55.55 & 12.58 & 4.89 \\
& \textbf{BEACON mean$\pm$SD} & \textbf{73.33$\pm$.31} & 55.43 & 12.68 & 5.22 \\
\bottomrule
\end{tabular}
\end{table}

\subsection{Selection and Test-Use Audit}
\label{sec:appx-selection}
Tab.~\ref{tab:appx-selection} makes the selection boundary explicit. No test segmentation metric or downstream WER selects a checkpoint or decoder setting.

\begin{table}[H]
\centering
\scriptsize
\setlength{\tabcolsep}{2pt}
\caption{Training, selection, and test-use separation for the reported systems.}
\label{tab:appx-selection}
\begin{tabular}{@{}>{\raggedright\arraybackslash}p{.27\columnwidth}>{\raggedright\arraybackslash}p{.42\columnwidth}>{\raggedright\arraybackslash}p{.24\columnwidth}@{}}
\toprule
\textbf{Item} & \textbf{Chosen by} & \textbf{Test Set Role} \\
\midrule
Teacher checkpoint & validation mF1B & final reporting only \\
Student checkpoint & validation balanced score & final reporting only \\
IO refinement $(\alpha,r,\tau)$ & validation grid & final reporting only \\
B/E fusion and decoder biases & fixed across primary runs & final reporting only \\
Architecture controls & validation balanced score & final reporting only \\
Renz et al. checkpoint & released artifact & evaluation only \\
Hands-On checkpoints & seed 42 recovered; seeds 43/44 fixed epoch 40 & evaluation only \\
\bottomrule
\end{tabular}
\end{table}

\subsection{Distillation Component Diagnostic}
\label{sec:appx-distill}
The non-distilled and distilled students use the same 5M-parameter architecture and strict no-cut endpoint policy.
Across three matched seeds, distillation changes mean \mFoneB/\mFoneS/\wer{} from 71.55/65.22/40.72 to 71.93/65.70/40.63 on PHOENIX-14 and from 68.10/58.17/73.74 to 68.50/58.42/73.33 on the BSL Corpus.
Tab.~\ref{tab:appx-distill} reports every paired run.

\begin{table}[H]
\centering
\scriptsize
\setlength{\tabcolsep}{3.8pt}
\caption{Matched distillation ablation under strict no-cut evaluation. The non-distilled and distilled rows use the same shared-student architecture and readout.}
\label{tab:appx-distill}
\begin{tabular}{llrrrr}
\toprule
\textbf{Corpus} & \textbf{Training} & \textbf{Seed} & \textbf{mF1B$\uparrow$} & \textbf{mF1S$\uparrow$} & \textbf{WER$\downarrow$} \\
\midrule
\multirow{6}{*}{PHX}
& No KD & 42 & 71.74 & 65.26 & 40.57 \\
& No KD & 43 & 71.25 & 65.05 & 40.68 \\
& No KD & 44 & 71.65 & 65.34 & 40.92 \\
& KD & 42 & 71.80 & 65.54 & 40.79 \\
& KD & 43 & 71.85 & 65.83 & 40.41 \\
& KD & 44 & 72.15 & 65.74 & 40.69 \\
\midrule
\multirow{6}{*}{BSL}
& No KD & 42 & 67.97 & 57.88 & 73.80 \\
& No KD & 43 & 68.35 & 58.37 & 74.42 \\
& No KD & 44 & 67.99 & 58.26 & 72.99 \\
& KD & 42 & 67.92 & 57.39 & 73.33 \\
& KD & 43 & 69.18 & 58.42 & 73.64 \\
& KD & 44 & 68.41 & 59.45 & 73.02 \\
\bottomrule
\end{tabular}
\end{table}

\subsection{Endpoint-Cut Compatibility Check}
\label{sec:appx-cut}
The endpoint-cutting post-processing step in Renz et al. reads the ground-truth start/end extent of each evaluation video.
It is therefore inappropriate for deployable headline inference, but paired evaluation quantifies its effect (see Tab.~\ref{tab:appx-cut}).
The official paper reports multi-run cut aggregates of $71.50/52.78$ on PHOENIX-14 and $68.68/47.71$ on the BSL Corpus, but only one linked checkpoint per corpus was available to us.
We therefore do not present our paired values as exact no-cut conversions of the published means.

\begin{table}[H]
\centering
\scriptsize
\caption{Paired no-cut/cut results from identical predictions. \bioecvr is a three-seed mean; Renz et al. is one checkpoint.}
\label{tab:appx-cut}
\begin{tabular}{llrrr}
\toprule
\textbf{Corpus} & \textbf{Segmenter/Protocol} & \textbf{mF1B$\uparrow$} & \textbf{mF1S$\uparrow$} & \textbf{WER$\downarrow$} \\
\midrule
\multirow{4}{*}{PHX}
& Renz et al./no-cut & 69.76 & 51.12 & 49.73 \\
& Renz et al./cut & 70.28 & 51.66 & --- \\
& BEACON/no-cut & 71.93 & 65.70 & 40.63 \\
& BEACON/cut & 72.08 & 65.78 & 40.58 \\
\midrule
\multirow{4}{*}{BSL}
& Renz et al./no-cut & 67.28 & 43.83 & 77.48 \\
& Renz et al./cut & 68.32 & 45.00 & --- \\
& BEACON/no-cut & 68.50 & 58.42 & 73.33 \\
& BEACON/cut & 68.87 & 58.66 & 73.17 \\
\bottomrule
\end{tabular}
\end{table}

\subsection{Fair Labeling and Decoder Factorial}
\label{sec:appx-fair}
The factorial comparison fixes the 1024-D input, 5-block TCN plus 4-layer Transformer, optimizer, losses, 50 epochs, split, paired seeds, selection rule, and no-cut evaluator.
BIO and BIOE differ by 257 output-head parameters (0.0058\%); no IO head, teacher, distillation, Begin/End auxiliary loss, or relocation is used.
Each constrained row uses the grammar matching its label representation: the BIO grammar disallows an \texttt{I} start or \texttt{O}$\to$\texttt{I}, while BIOE uses the declared eight-edge grammar.
Paired BIOE-minus-BIO differences under these matching grammars are $-0.30\pm0.39/-0.04\pm0.62/-0.53\pm1.16$ (\mFoneB/\mFoneS/WER) on PHOENIX-14 and $+1.50\pm0.46/+0.92\pm0.77/0.00\pm0.88$ on the BSL Corpus (see Tab.~\ref{tab:appx-fair}).
With three paired seeds, exact sign-flip tests have little resolution; we therefore emphasize effect estimates and direction consistency rather than significance.

\begin{table}[H]
\centering
\scriptsize
\setlength{\tabcolsep}{3.5pt}
\caption{Fair ablation, three-seed mean $\pm$ SD. WER is reported for grammar-constrained readout only.}
\label{tab:appx-fair}
\begin{tabular}{lllrrr}
\toprule
\textbf{Corpus} & \textbf{Labels} & \textbf{Readout} & \textbf{mF1B$\uparrow$} & \textbf{mF1S$\uparrow$} & \textbf{WER$\downarrow$} \\
\midrule
\multirow{4}{*}{PHX} & BIO & argmax & 70.78$\pm$.14 & 62.26$\pm$.19 & --- \\
 & BIO & grammar & 70.68$\pm$.16 & 62.80$\pm$.37 & 43.82$\pm$.15 \\
 & BIOE & argmax & 70.56$\pm$.34 & 62.25$\pm$.41 & --- \\
 & BIOE & 8-edge & 70.38$\pm$.51 & 62.76$\pm$.27 & 43.29$\pm$1.01 \\
\midrule
\multirow{4}{*}{BSL} & BIO & argmax & 64.79$\pm$.58 & 54.85$\pm$.58 & --- \\
 & BIO & grammar & 63.98$\pm$.55 & 54.54$\pm$.75 & 73.95$\pm$.25 \\
 & BIOE & argmax & 66.27$\pm$.24 & 55.43$\pm$.29 & --- \\
 & BIOE & 8-edge & 65.48$\pm$.09 & 55.45$\pm$.03 & 73.95$\pm$.67 \\
\bottomrule
\end{tabular}
\end{table}

\subsection{Learned-Transition CRF}
\label{sec:appx-crf}
The learned transition control uses the same BIOE encoder and training split but an unrestricted linear-chain CRF.
We count a decoded sequence as invalid when it contains any start, end, or adjacent-state transition outside the fixed eight-edge grammar.
The CRF is competitive, but learned scores do not enforce the known grammar.
Constrained decoding always returns a path in that grammar and adds no learned transition parameters.

\begin{table}[H]
\centering
\scriptsize
\setlength{\tabcolsep}{2.5pt}
\caption{Learned transition CRF, strict no-cut, three-seed mean $\pm$ SD. Invalid is the mean percentage of test sequences containing at least one transition outside the fixed grammar.}
\label{tab:appx-crf}
\begin{tabular}{lrrrr}
\toprule
\textbf{Corpus} & \textbf{mF1B$\uparrow$} & \textbf{mF1S$\uparrow$} & \textbf{WER$\downarrow$} & \textbf{Invalid (\%)} \\
\midrule
PHX & 70.13$\pm$1.29 & 62.78$\pm$2.09 & 43.77$\pm$2.75 & 12.65 \\
BSL & 66.40$\pm$1.06 & 56.53$\pm$1.45 & 73.43$\pm$1.15 & 12.63 \\
\bottomrule
\end{tabular}
\end{table}

\subsection{Architecture Families Metric--WER Analysis}
\label{sec:appx-family}
Tab.~\ref{tab:appx-families} retains sample-SD variation for three-seed models and reports no artificial variance for the public Renz et al. checkpoint.
Exact permutation tests are used because $n=5$.
The Spearman $(\rho,p)$ pairs for \mFoneB/\mFoneS{} are $(-.90,.083)/(-1.00,.017)$ on PHOENIX-14 and $(-.80,.133)/(-.90,.083)$ on the BSL Corpus.
Leave-one-family-out ranges are $[-1.00,-.80]/[-1.00,-1.00]$ and $[-.80,-.60]/[-1.00,-.80]$, respectively.
These sensitivity results reinforce the ordering observation but not a universal claim.

\begin{table}[H]
\centering
\scriptsize
\caption{Architecture families comparison (three-seed mean $\pm$ SD except the single public Renz et al. checkpoint). Local Transformer is an architecture control, not an exact ASFormer reproduction. Exact permutation $p$-values accompany Spearman $\rho$.}
\label{tab:appx-families}
\begin{tabular}{lccc}
\toprule
\textbf{Family} & \textbf{mF1B$\uparrow$} & \textbf{mF1S$\uparrow$} & \textbf{WER$\downarrow$} \\
\midrule
\multicolumn{4}{l}{\textit{PHX}} \\
MS-TCN & 69.76 & 51.12 & 49.73 \\
Local Transformer & 48.29$\pm$.54 & 37.03$\pm$.07 & 50.90$\pm$.92 \\
BiLSTM-CRF & 67.80$\pm$.10 & 58.08$\pm$.26 & 47.15$\pm$.75 \\
MS-TCN++ & 70.60$\pm$.67 & 60.10$\pm$1.71 & 42.54$\pm$.29 \\
BEACON & \textbf{71.93$\pm$.19} & \textbf{65.70$\pm$.15} & \textbf{40.63$\pm$.20} \\
\midrule
$\rho$ with WER ($p$) & $-.90$ (.083) & $-1.00$ (.017) & --- \\
\midrule
\multicolumn{4}{l}{\textit{BSL}} \\
MS-TCN & 67.28 & 43.83 & 77.48 \\
Local Transformer & 49.97$\pm$1.16 & 36.36$\pm$.84 & 79.41$\pm$.49 \\
BiLSTM-CRF & 62.59$\pm$.26 & 50.94$\pm$.15 & 76.15$\pm$.13 \\
MS-TCN++ & 67.68$\pm$1.33 & 55.78$\pm$.93 & \textbf{73.12$\pm$.63} \\
BEACON & \textbf{68.50$\pm$.64} & \textbf{58.42$\pm$1.03} & 73.33$\pm$.31 \\
\midrule
$\rho$ with WER ($p$) & $-.80$ (.133) & $-.90$ (.083) & --- \\
\bottomrule
\end{tabular}
\end{table}

\subsection{Inference Efficiency}
\label{sec:appx-efficiency}
We measured the neural forward pass on an NVIDIA RTX 5090, FP32, batch size 1, at the p50/p90 test lengths of each corpus, with 20 warm-up and 100 CUDA-synchronized repetitions.
CPU Viterbi and endpoint relocation are excluded from both systems.
The shared student reduces parameters by $43.3\%$, resident allocation by $43.5\%$, peak allocation by approximately $29\%$--$33\%$, and forward latency by $36.5\%$--$36.9\%$.

\begin{table}[H]
\centering
\scriptsize
\setlength{\tabcolsep}{3.5pt}
\caption{Inference efficiency. Latency ranges cover both corpora and sequence lengths.}
\label{tab:appx-efficiency}
\begin{tabular}{lcccc}
\toprule
\textbf{System} & \textbf{Params} & \textbf{Resident} & \textbf{Peak} & \textbf{Latency} \\
& & \textbf{(MiB)} & \textbf{(MiB)} & \textbf{(ms)} \\
\midrule
Two encoders & 8.82M & 34.65 & 45.25--47.71 & 1.41--1.43 \\
Single distilled & 5.00M & 19.59 & 30.19--32.65 & 0.89--0.90 \\
\bottomrule
\end{tabular}
\end{table}

\subsection{Reproducibility Notes}
All 12 fair ablation runs, 18 architecture control runs, 6 teachers, 6 non-distilled students, 6 distilled students, and 6 learned CRF runs completed without NaN, traceback, or out-of-memory failure.
The fixed split has zero video overlap and covers the original training set exactly.
Teacher and student use identical indices; teachers are frozen and excluded from the inference checkpoint.
Corpus WER denominators are constant across segmenters ($N=10{,}987$ PHOENIX-14; $N=4{,}214$ BSL Corpus).
The split is not recording group-disjoint because most available recording groups occur on both sides of the original training pool; this is disclosed as a limitation rather than treated as a cross-group validation result.
Code and evaluation scripts are included in the Supplementary archive and will be released publicly upon acceptance.

\subsection{Data and Ethical Scope}
The experiments use previously released sign language research corpora and derived features under their existing access conditions.
We do not infer signer identity or demographic attributes.
Automatic lexical segmentation can support annotation and retrieval, but errors may disproportionately affect signing styles underrepresented in the training corpora; the present two-corpus evaluation does not quantify that disparity.
Predicted boundaries and recognition outputs should therefore not be treated as authoritative linguistic annotation without human review.

\end{document}